\documentclass[10pt,twocolumn,letterpaper]{article}

\usepackage{cvpr}
\usepackage{times}
\usepackage{epsfig}
\usepackage{graphicx}
\usepackage{amsmath}
\usepackage{amssymb}
\usepackage{booktabs}
\usepackage{makecell}
\usepackage{algpseudocode}
\usepackage{cite}

\usepackage{float}

\usepackage[pagebackref=true,breaklinks=true,letterpaper=true,colorlinks,bookmarks=false]{hyperref}

\def\muprior{\mu_\text{prior}}
\def\sigmaprior{\sigma_\text{prior}}
\def\sigmapriorSquared{\sigma_\text{prior}^2}
\DeclareMathOperator*{\median}{median}
\DeclareMathOperator*{\argmin}{argmin}
\newcommand{\uniform}[2]{\operatorname{unif}(#1, #2)}

\cvprfinalcopy %

\def\cvprPaperID{7} %

\ifcvprfinal\fi

\begin{document}

\title{Minimizing Supervision for Free-space Segmentation}

\author{
Satoshi Tsutsui\thanks{%
Part of this work was done as an intern at Preferred Networks, Inc.
} \textsuperscript{, }\footnotemark[2]\\
Indiana University\\
{\tt\small stsutsui@indiana.edu}
\and
Tommi Kerola \footnotemark[2] \hspace{10mm} Shunta Saito
\thanks{The first three authors contributed equally. The order was decided by using the paper ID as a random seed and then calling \texttt{np.random.permutation}.}\\
Preferred Networks, Inc.\\
{\tt\small \{tommi,shunta\}@preferred.jp}
\and
David J. Crandall\\
Indiana University\\
{\tt\small djcran@indiana.edu}
}

\maketitle

\begin{figure*}[t!]
\setlength{\linewidth}{\textwidth}
\setlength{\hsize}{\textwidth}
  \centering
  \includegraphics[width=\linewidth]{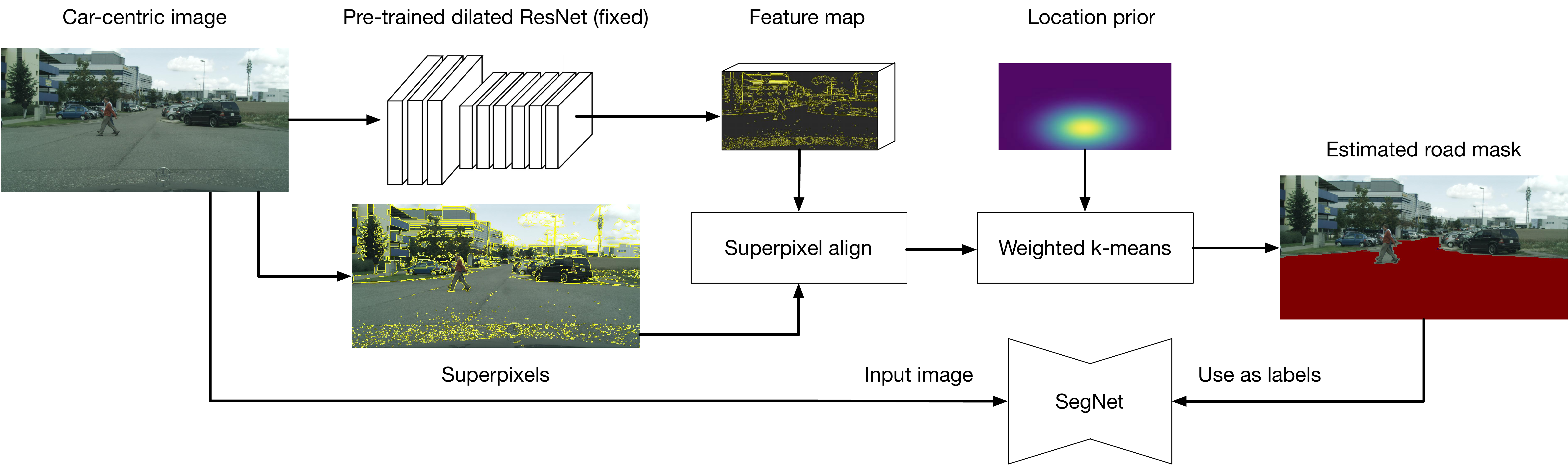}
 	\caption{Overview of our method. We extract features from a dilated ResNet pre-trained on ImageNet and leverage a novel superpixel alignment method
 	and location prior clustering to generate masks for training a segmentation CNN. Our method requires no manual annotation of free-space labels. Best viewed in color.}
	\label{fig:method_overview}
\end{figure*}

\begin{abstract}
Identifying ``free-space,'' or safely driveable regions in the scene
ahead, is a fundamental task for autonomous navigation.  While this
task can be addressed using semantic segmentation, the manual labor involved in
creating pixel-wise annotations to train the segmentation model
is very costly. Although weakly supervised segmentation addresses this
issue, most methods are not designed for free-space. In this paper, we
observe that homogeneous texture and location are two key characteristics
of free-space, and develop a novel, practical
framework for free-space segmentation with minimal human
supervision. Our experiments show that our framework performs better
than other weakly supervised methods while using less
supervision. Our work demonstrates the potential for performing free-space
segmentation without tedious and costly manual annotation, which will
be important for adapting autonomous driving systems to different types of
vehicles and environments.

\end{abstract}

\section{Introduction}

A critical perceptual problem in autonomous vehicle navigation is
deciding whether the path ahead is safe and free of potential
collisions.  While some problems (like traffic sign detection) may
just require detecting and recognizing objects, avoiding collisions
requires fine-grained, pixel-level understanding of the scene in front
of the vehicle, to separate ``free-space''~\cite{survey} --
road surfaces that are free of obstacles, in the case of autonomous cars, for example -- from other scene
content in view.

Free-space segmentation can be addressed by existing fully-supervised
semantic segmentation algorithms~\cite{oliveira2016efficient}. But a
major challenge is the cost of obtaining pixel-wise ground truth
annotations to train these algorithms: human-labeling of a single object in a
single image 
can take approximately 80 seconds~\cite{Bearman2016}, while
annotating all road-related objects in a street scene
may take over an hour~\cite{cordts2016cityscapes}. The high cost of collecting training data may be a substantial barrier for developing autonomous driving systems for new environments that have not yet received commercial attention (e.g.\ in resource-poor countries, for off-road contexts, for autonomous water vehicles, etc.), and especially for small companies and research groups with limited resources.

In this paper, we develop a framework for free-space segmentation that
minimizes human supervision.  Our approach is based on two
straightforward observations. First, free-space has a strong
\textit{location prior}:  pixels corresponding to free
space are likely to be located at the bottom and center of the image
taken by a front-facing camera, since in training data there  is
always free-space under the vehicle (by definition).  Second, a free-space
region generally has \textit{homogeneous texture} since road
surfaces are typically level and smooth (e.g. concrete
or asphalt in an urban street).  

To take advantage of these observations, we first group together
pixels with low-level homogeneous texture into superpixels. We then
select candidate free-space superpixels through a simple clustering
algorithm that incorporates both the spatial prior and appearance
features (\S\ref{method:clustering}).  The remaining challenge is to
create higher-level features for each superpixel that semantically
distinguish free-space.  We show that features from a CNN pre-trained
on ImageNet~(\S\ref{method:features})
perform 
well for free-space when combined with \emph{superpixel alignment}, a  novel method that aligns superpixels with CNN feature
maps~(\S\ref{method:superpixel_align}). Finally, these results are
used as labels to train a supervised segmentation
method~(\S\ref{method:training}) for performing segmentation on new
images.

We note that our framework does not need any image annotations, so
collecting annotated data is a simple matter of recording
vehicle-centric images while navigating the environment where
free-space segmentation is needed, and then running our algorithm. The
human effort required is reduced to specifying the location prior and
adjusting hyper-parameters such as superpixel granularity and the
number of clusters. This form of supervision requires little effort
because the technique is not very sensitive to the exact values of
these parameters, as we empirically demonstrate with experiments on the
well-established, publicly-available Cityscapes dataset~\cite{cordts2016cityscapes}. Our
quantitative evaluation shows that our framework yields better
performance than various baselines, even those that use more
supervision than we do~(\S\ref{experiments}) .

In summary, we make the following contributions:
\begin{itemize}
    \setlength{\parskip}{0cm} %
    \setlength{\itemsep}{0cm} %
    \item We develop a novel framework for free-space segmentation
      that does not require any image-level annotations, by taking
      advantage of the unique characteristics of free-space;
    \item We propose a novel algorithm for combining CNN feature maps
      and superpixels, and a clustering method that incorporates 
      prior knowledge about the location of free-space; and
    \item We show that our approach performs better than other
      baselines, even those that require more supervision.
\end{itemize}

\section{Related Work}
\paragraph{Fully supervised segmentation.}
Many recent advances in semantic segmentation have been built on fully
convolutional networks (FCNs)~\cite{long2015fully}, which extend
CNNs designed for image classification by posing semantic segmentation
as a dense pixel-wise classification problem.  This dense
classification requires high resolution feature maps for prediction, so
FCNs add upsampling layers into the classification CNNs (which otherwise usually
perform downsampling through pooling
layers). SegNet~\cite{BadrinarayananSegnet} improves upon this and
introduces an unpooling layer for upsampling, which reflects the
pooling indices used in the downsampling phase. We use SegNet
here, although our technique is flexible enough to be used with other
FCNs as well.

A problem with CNN pooling layers is that they discard spatial
information that is critical for image segmentation. One solution is to use dilated (or `atrous')
convolutions~\cite{yu2015multi}, which allow receptive field expansion
without pooling layers. Dilated convolutions have been incorporated
into recent frameworks such as DeepLab~\cite{chen2016deeplab} and
PSPNet~\cite{zhao2016pyramid}.  Although our work does not focus on
engineering CNN architectures, this direction inspired our choice of
CNN for image feature extraction, since we similarly want to obtain a
high resolution feature map. In particular, we use dilated
ResNet~\cite{Yu_2017_CVPR} trained on ImageNet, yielding a higher
resolution feature map than the normal ResNet~\cite{he2016deep}.

\paragraph{Weakly supervised segmentation.}
Since ground-truth segmentation annotations are very costly to obtain,
many techniques for segmentation have been proposed
that require weaker annotations, such as image
tags~\cite{pinheiro2015image,kolesnikov2016seed,shimoda2016distinct,Saleh2017IncorporatingNB,Durand2017,wei2017adverserial},
scribbles~\cite{Lin2016}, bounding boxes~\cite{Khoreva2017}, or videos
of objects~\cite{Tokmakov16a,Hong2017WeaklySS}. At a high level of
abstraction, our work can be viewed as a tag-based weakly supervised
method, in that we assume all images have a ``tag'' of
free-space. However, most previous studies mainly focus on foreground
objects, so are not directly applicable for free-space, which can be
regarded as background~\cite{Saleh_2017_ICCV}. From a technical perspective, some
methods propose new CNN architectures~\cite{Durand2017} or better
loss functions~\cite{kolesnikov2016seed}, while others focus on
automatically generating segmentation masks for training available
CNNs~\cite{Khoreva2017}. We follow the latter approach here of generating segmentation
masks for CNNs. We also do not use the approach of gradually refining
the segmentation mask~\cite{shimoda2016distinct}, because we believe
that autonomous vehicles require a high-quality trained CNN even at the
stage of initial deployment.

\paragraph{Free-space segmentation.}
 Free-space segmentation is the task of estimating the space through which a
 vehicle can drive safely without collision. This task is critical for
 autonomous driving and has traditionally been addressed by geometric
 modeling~\cite{kong2009vanishing,alvarez20103d,badino2007free,wedel2009b},
 handcrafted features~\cite{alvarez2011road,hanisch2017free}, or even a
 patch-based CNN~\cite{alvarez2012road}.  We use FCNs in this
 paper, which Oliveira~\etal~\cite{oliveira2016efficient} demonstrated
 to be efficient for road segmentation.  

Since pixel-wise ground truth annotations are so expensive to obtain,
several papers have investigated weakly supervised free-space
segmentation. While an early study~\cite{guo2012robust} trains a
probabilistic model, other papers train
FCNs~\cite{Tsutsui_2017_ICCV_Workshops,Saleh_2017_ICCV,laddha2016map,Sanberg2017EIAVM}.
Saleh~\etal~\cite{Saleh_2017_ICCV} develop a video segmentation
algorithm for general background objects including free-space on a 
road. Tsutsui~\etal~\cite{Tsutsui_2017_ICCV_Workshops} propose
distantly supervised monocular image segmentation. However, both
methods require additional images to train a saliency or attention
extractor. Laddha~\etal~\cite{laddha2016map} use external maps of the
road indexed against the vehicle position according to GPS.
Sanberg~\etal~\cite{Sanberg2017EIAVM} and
Guo~\etal~\cite{guo2012robust} use stereo information for
automatically generating segmentation masks. We distinguish our work
from these studies in that we only use a collection of
monocular vehicle-centric images, which makes our approach even less
supervised than most  others.

For evaluating free-space segmentation,
KITTI~\cite{Geiger2012AreWR} and CamVid~\cite{brostow2009semantic}
are older datasets that 
are not large enough to leverage the
power of CNNs. Recently, a larger dataset called
Cityscapes~\cite{cordts2016cityscapes} was proposed for object
segmentation in autonomous
driving. 
We conduct our experiments on Cityscapes, since existing work has
found that CNNs trained on Cityscapes perform better than other state-of-the-art methods on KITTI and CamVid~\cite{cordts2016cityscapes}.

\section{Our approach}\label{sec:method}

We now describe our technique for automatically generating annotations
suitable for training a free-space segmentation CNN. 
Our technique relies on two main assumptions about the nature of
free-space: (1) that free-space regions tend to have homogeneous
texture (e.g., caused by smooth road surfaces), and (2) there are strong priors on
the location of free-space within an image taken from a vehicle.  The
first assumption allows us to use superpixels to group 
similar pixels. As in previous
work~\cite{Tsutsui_2017_ICCV_Workshops,Lin2016,Tokmakov16a,Hong2017WeaklySS,chen2017no},
we use the Felzenszwalb and Huttenlocher graph-based segmentation
algorithm~\cite{felzenszwalb2004efficient} to create the superpixels,
since the specific superpixel algorithm is not the focus of this
study.

The second assumption allows us to find ``seed'' superpixels
that are very likely to be free-space, based on the fact that free-space
is usually near the bottom and center of an image taken by
a front-facing in-vehicle camera.
A very naive method would be to
select superpixels covering predefined locations based
on the prior, but this would  ignore any semantic or higher level
features other than the texture features used for
generating superpixels. We thus cluster superpixels based on semantic
features and automatically select the cluster likely corresponding to
free-space based on the location prior, as described in \S\ref{method:clustering}.
We perform this clustering on multiple images at a time, to be robust
against occasional images which do not satisfy the prior assumption.

An important question is how to extract semantic-level features from
each superpixel.  We show that the features extracted from 
CNNs pre-trained on ImageNet are generic enough~(\S\ref{method:features}) for our task, 
and 
we develop a novel technique called \textit{superpixel
  alignment} that efficiently aggregates CNN features for the region within a
superpixel~(\S\ref{method:superpixel_align}). Finally, superpixel
clustering automatically generates a free-space pixel mask, which we
then use to train supervised CNNs for
segmentation~(\S\ref{method:training}) \footnote{The source code for automatic mask generation is available at\\ { \url{https://github.com/apple2373/min-seg-road} } }.

\begin{figure}[t!]
  \centering
  \includegraphics[width=\linewidth]{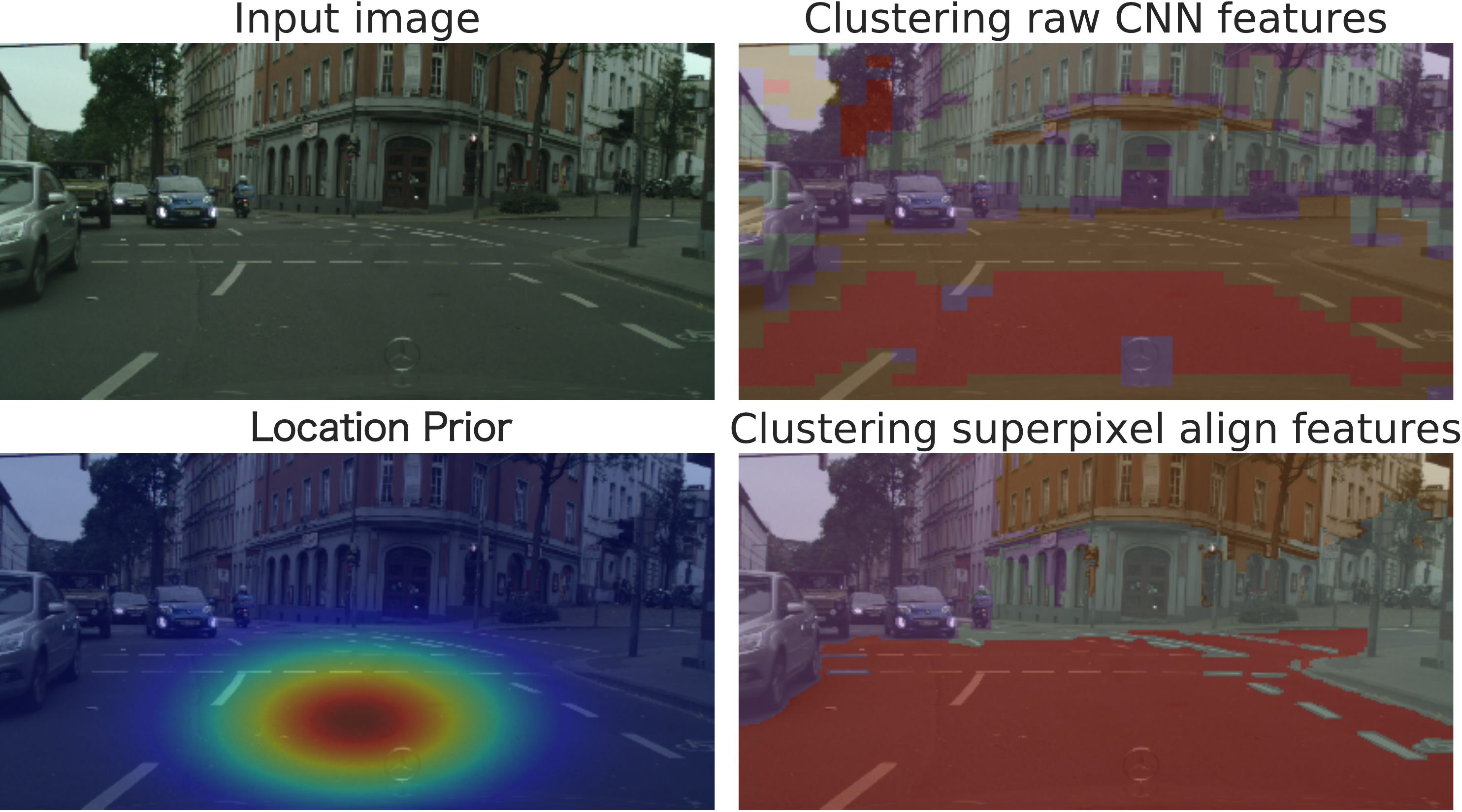}
 	\caption{Example of using location prior k-means on dilated ResNet (CNN)     or superpixel alignment features to find the location of the free-space.
 	Note that due to the design of our algorithm, the free-space will typically be the first cluster (red),
 	which means that we do not need any sophisticated post-processing in order to find the free-space cluster.
 	Best viewed in color.}
	\label{fig:weighted_kmeans_example}
\end{figure}

The reader may wonder why we do not cluster the CNN features directly,
given that they capture semantic information. However, because certain
parts of free-space is semantically more important than others, direct clustering results would not be smooth and
cohesive. This is visually confirmed by comparing the two clustering
results shown in Figure~\ref{fig:weighted_kmeans_example}.

\subsection{Features for Clustering}\label{method:features}
We cluster superpixels based on features extracted from a CNN
pre-trained on ImageNet.  Such features have been found to capture
latent features having rich semantic information for semantic
segmentation~\cite{Saleh2017IncorporatingNB}, even though the ImageNet
challenge does not include free-space as one of its annotated classes.
 Much work has found that these features are surprisingly
general across vastly different domains including document image
analysis~\cite{harley2015evaluation} and medical image
analysis~\cite{esteva2017dermatologist}.  This is probably because
early layers in convolutional neural networks tend to learn low-level
features (\eg, edges), while later layers capture increasing amounts of
semantic information, with the final layers capturing features
suitable for the explicit classification problem (\eg, object types
like cars) that the network was trained to
solve~\cite{Saleh2017IncorporatingNB}. We confirmed this tendency by visualizing feature maps of the 26-layer dilated ResNet~\cite{Yu_2017_CVPR} that is
 trained for the task of ImageNet classification, and decided to use the last layer feature map, which indeed seemed to capture higher level information. We note that this type of manual inspection has also been performed in previous work~\cite{Saleh2017IncorporatingNB}.

Among other CNN architectures, we
intentionally select a dilated network architecture in order to produce
higher resolution feature maps, which are important for being
able to localize the road with a fine level of granularity.

\begin{figure}[t]
\scalebox{0.9}{
\begin{minipage}{1.1\linewidth}
\begin{algorithmic}
\Function{locationPriorKMeans}{$k$, $\{S_i\}_{\forall i}$, $\muprior{}$, $\sigmaprior{}$}
\State $\forall i, \forall S_{xy} \in S_i: p_{xy} = \operatorname{spatial\_coord}(S_{xy})$
\State $\forall i: w_i = \frac{1}{|S_i|} \sum_{S_{xy} \in S_i} e^{-\|p_{xy}-\muprior{}\|^2 / (2\sigmapriorSquared{})}$
\State $\forall i: m_i \gets 
    \begin{cases}
    0 & \text{if } w_i > \median_{j}{w_j} \\
    \uniform{1}{k-1} & \text{otherwise}
    \end{cases}$
\While{not converged}
    \State $c_0 \gets \sum_{m_i = 0} w_i S_i / \sum_{m_i=0} w_i$
    \State $\forall q>0: c_q \gets \sum_{m_i = q} (1-w_i) S_i / \sum_{m_i=q} (1-w_i)$
    \State $\forall i: m_i \gets \argmin_{q} \| S_i - c_q \|_2^2$
\EndWhile
    \State \Return $\{ c_0, \ldots, c_{k-1}\}$ \Comment{Cluster centers}
\EndFunction
\end{algorithmic}
\end{minipage}
}
\vspace{1mm}
\caption{Location prior k-means.}
\label{fig:weighted_kmeans}
\end{figure}

\begin{figure}[t!]
  \centering
  \includegraphics[width=\linewidth]{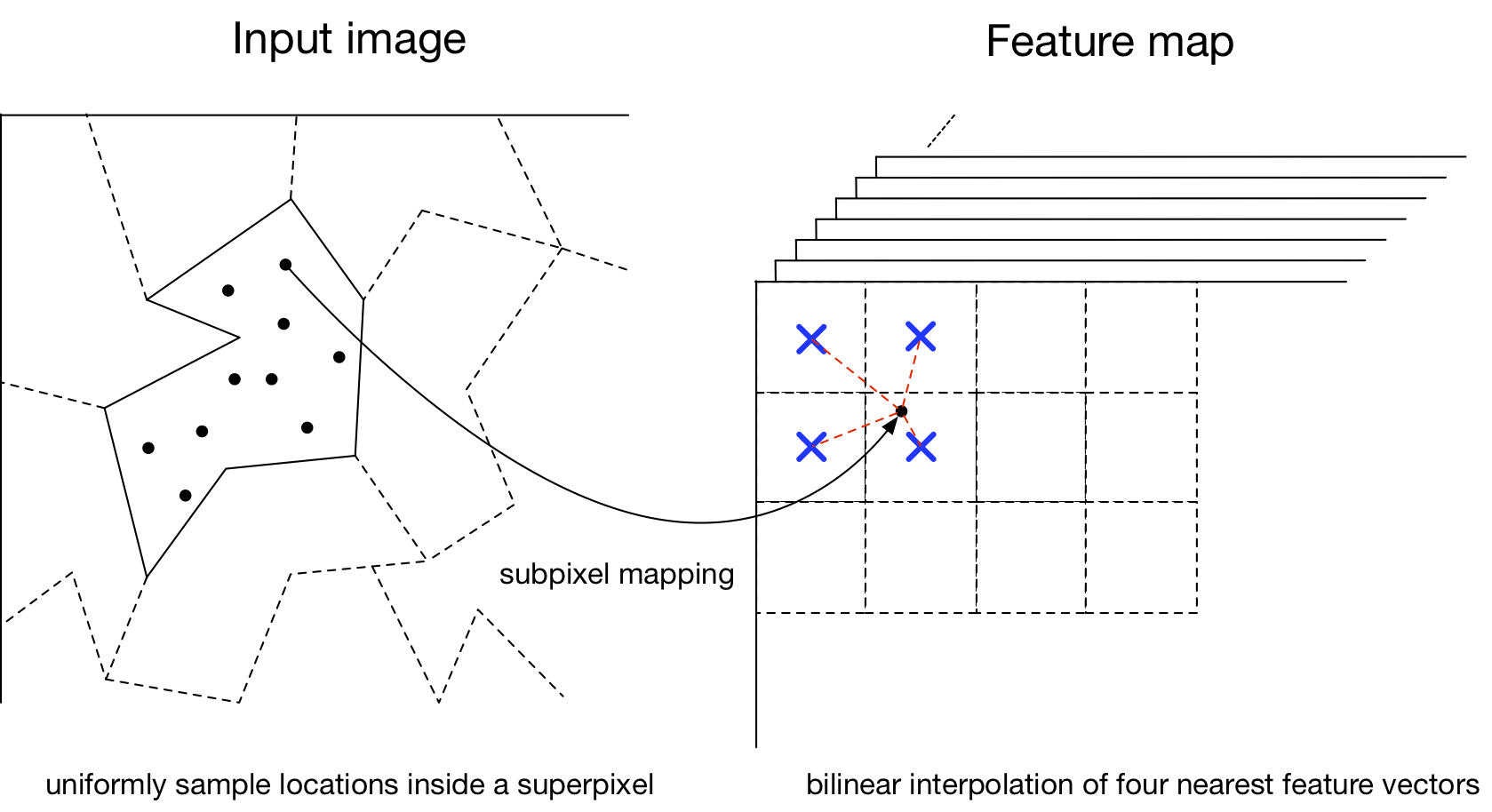}
 	\caption{Illustration of superpixel alignment. The feature
          vector for each superpixel is defined by randomly sampling
          10 points inside the superpixel, and taking the average of
          their bilinearly interpolated feature maps.}
	\label{fig:superpixel_align}
\end{figure}

\subsection{Superpixel Alignment}\label{method:superpixel_align}

We now wish to extract appearance features for each superpixel.  While
the dilated ResNet features capture semantic information, they are not
well localized for free-space, so we align them to spatially
coherent superpixels to create a better representation of the scene.
To do this, we propose a new method called
\textit{superpixel alignment}, which is inspired by
RoIAlign~\cite{He_2017_ICCV}. The technique applies bilinear
interpolation of the CNN feature maps for a random subset of the pixels
inside each superpixel.  More precisely, we perform bilinear
interpolation~\cite{jaderberg2015spatial} of dilated ResNet features
$F_{mn}^c$ at spatial location $(m, n)$ and channel $c$ as
\begin{align}
    S_{xy}^c &= \sum_{(m,n) \in \mathcal{N}_{xy}} F_{nm}^c \max(0, \hat{\Delta}_{xm}) \max(0, \hat{\Delta}_{yn})~,
\end{align}
\noindent where $\hat{\Delta}_{xm} = 1 - |x - m|$ and
$\mathcal{N}_{xy}$ is the set of neighbors for spatial location $(x,
y)$ in superpixel $S$.  We sample 10 locations uniformly at random
inside each superpixel, and then use the four nearest neighbors of
each selected pixel for the bilinear interpolation. Finally, we
aggregate the features inside each superpixel using average pooling.
Note that unlike RoIAlign, we assume that each superpixel consists of a
homogeneous set of pixels; this  avoids the need for computing the bilinear
interpolation densely for all pixels by instead using a small randomly sampled
set, which we have found works
well in practice.  To improve the spatial cohesiveness of the feature,
we append the centroid of the spatial coordinates of
the superpixel to the pooled feature vector.  This gives us one image
feature for each superpixel.  The procedure for superpixel alignment is
summarized in Fig.~\ref{fig:superpixel_align}.

\begin{figure*}[t!]
  \centering
  \includegraphics[width=0.98\linewidth]{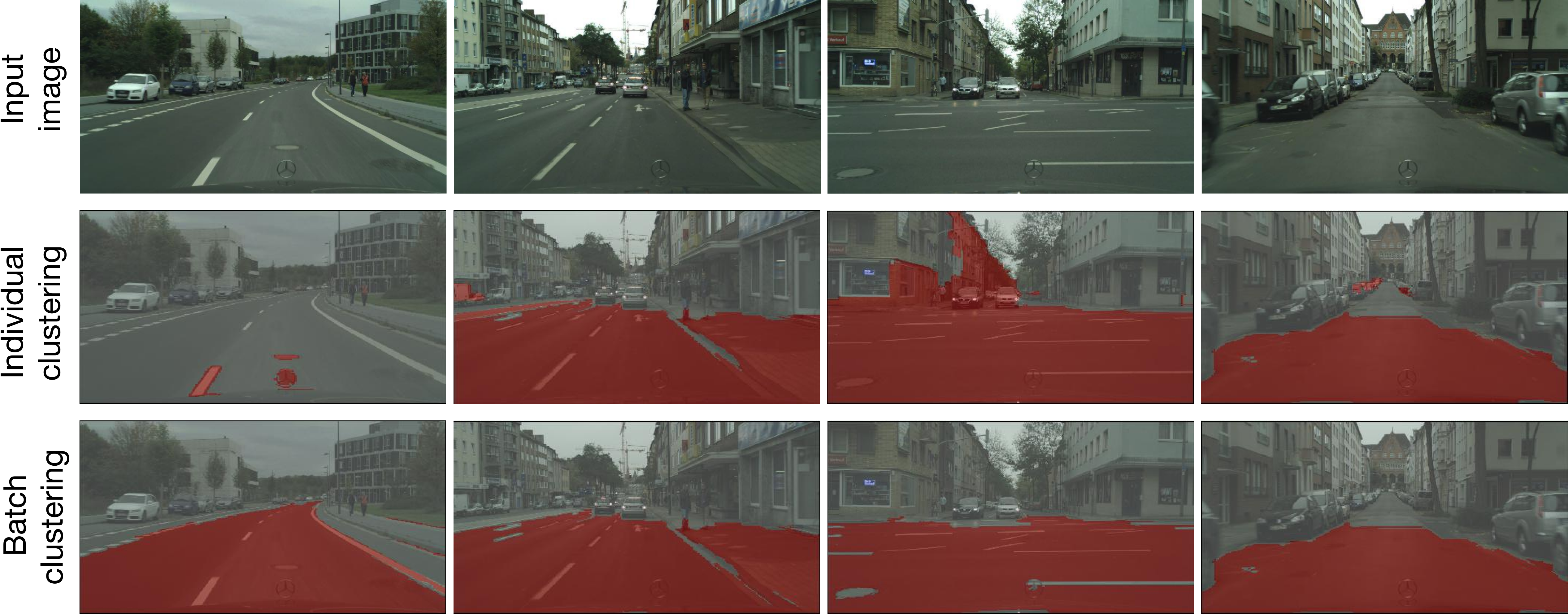}
 	\caption{Batch clustering. (top) Input images. (middle) Results of clustering each image individually. (bottom) Results of clustering the four images together in a batch manner. The left image has a spot around the center of the location prior so only the spot is recognized as free-space when clustered individually, but batch clustering avoids this mistake.}
	\label{fig:batch-clustering}
\end{figure*}

\subsection{Superpixel Clustering}
\label{method:clustering}

Using the features defined in the last section, we can now
apply any standard clustering algorithm.
An important remaining problem, however, is how
to determine which cluster corresponds with free-space.
A simple solution would be to select the largest cluster
appearing in the bottom half the image, for example, but this would
fail in crowded scenes with large numbers of foreground objects on the
road.

Instead, inspired by previous
work~\cite{tseng2007penalized,achanta2012slic}, we use
prior information about the spatial location of free-space, namely that the road
surface should usually be immediately above the visible chassis of the
ego-vehicle.  To do this, we
adapt Lloyd's algorithm~\cite{lloyd1982least} for solving a weighted
variant of the k-means clustering problem.  
We represent the prior as an average of Gaussians
\begin{align}
 w_i = \frac{1}{|S_i|} \sum_{S_{xy} \in S_i} \exp\left(-\frac{\|p_{xy}-\muprior{}\|^2 }{ 2\sigmapriorSquared{}}\right)~,
\end{align}
such that each superpixel $S_i$ has
a prior weight $w_i$ that is parameterized by $\muprior{}=[0.75, 0.5]$
and $\sigmaprior{}=[0.1,0.1]$ wrt. the image dimensions (estimated empirically) and the spatial
coordinates $p_{xy}$ of each pixel inside the superpixel.  In
practice, we manually adjust the prior parameters empirically with a 
small number of example images. Subsequently, we initialize half of
the pixels to the free-space cluster (which we assume is the first
cluster) based on these weights.  The first cluster is then encouraged
to consist of pixels corresponding to free-space by setting its
cluster center to be the spatially weighted average of features
assigned to it.  The other clusters have a repellent weight assigned
to their members to encourage them to spatially spread away from the
location prior.  Cluster memberships are updated in the same manner as
the standard k-means algorithm without taking the weights into
account.  Our algorithm is summarized in
Fig.~\ref{fig:weighted_kmeans}, and an example of the output of the
algorithm is shown in Fig.~\ref{fig:weighted_kmeans_example}.

Although our cluster update breaks the convergence criterion of
standard k-means clustering~\cite{christopher2008introduction}, we
have found that in practice it usually converges to a stable solution.  We
note that similar prior information could also be incorporated into
other types of clustering algorithms, such as spectral
clustering~\cite{shi2000normalized}.

\paragraph{Batch image clustering.} 
Of course, while the spatial prior assumption on free-space is
reasonable in general, it is often violated in individual images (e.g.,
a vehicle or pedestrian could be located in the center of the location prior,
which could cause the algorithm to incorrectly assign the first cluster to
consist of features corresponding to non-road locations). We circumvent this issue by clustering superpixels from multiple images at the same time, which we call batch clustering.  In Fig.~\ref{fig:batch-clustering}, we show an example where only a single spot at the center of the location prior is recognized as 
free-space, but clustering with three other images prevents this
mistake. As our experiments will show, batch clustering is effective
for generating higher quality segmentation masks.

\begin{figure}[t!]
  \centering
  \includegraphics[width=\linewidth]{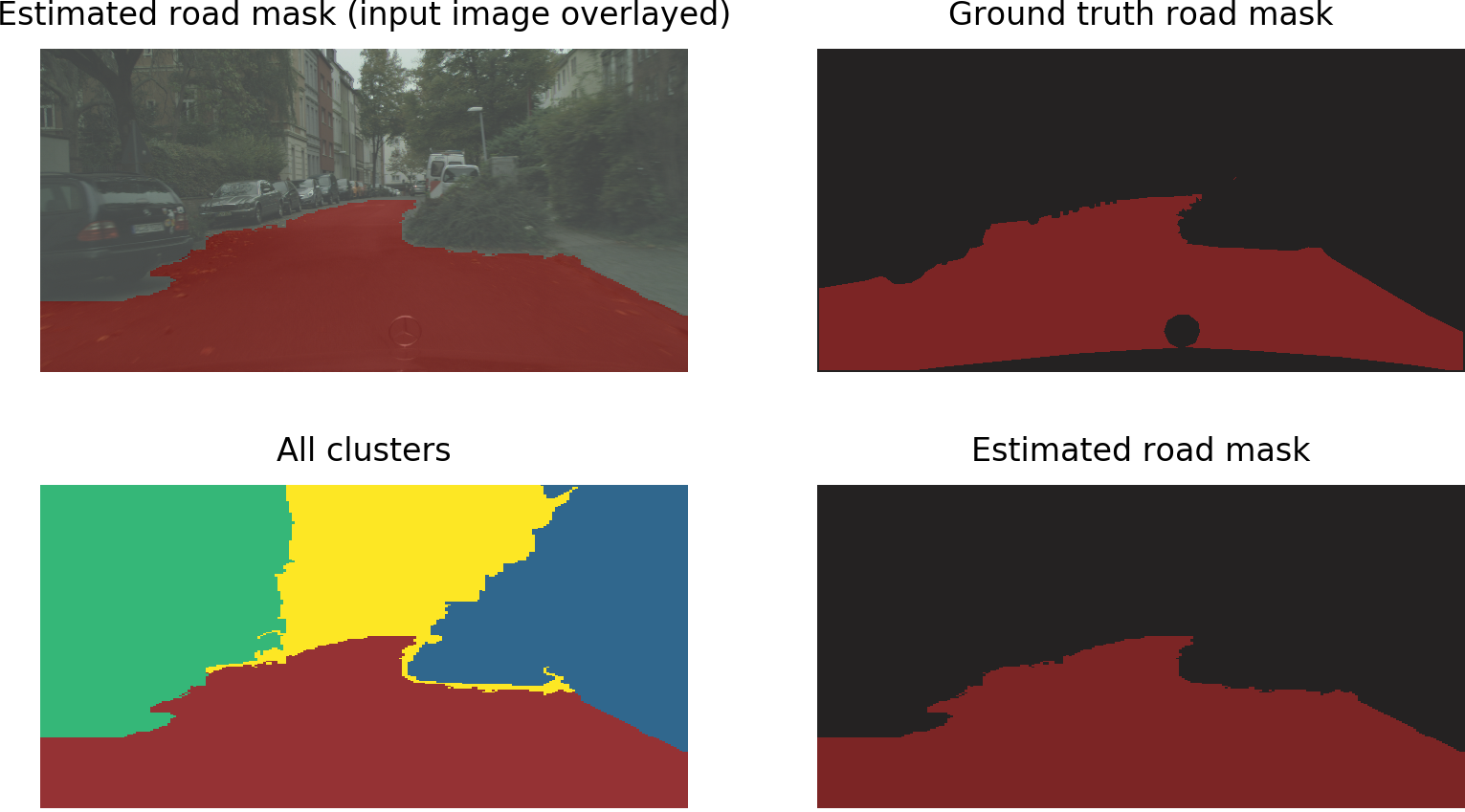}
 	\caption{Example of superpixel alignment compared to ground truth.}
	\label{fig:example_results}
\end{figure}

\subsection{CNN Training from Generated Mask}\label{method:training}
Once the masks for free-space have been obtained by superpixel
clustering (an example can be seen in Fig.~\ref{fig:example_results}),
we then use these automatically generated masks to train a road segmentation
CNN using supervised training.

\section{Experiments}\label{experiments}

\paragraph{Dataset.} 
We conducted a series of experiments on the established Cityscapes~\cite{cordts2016cityscapes} dataset to evaluate our proposed method.
This dataset is designed for evaluating  segmentation algorithms
for autonomous driving applications, and includes a set of
fine-grained pixel-wise annotations for 19 types of traffic objects.
We only use the `road' class and treat it as
free-space. We report the intersection over union (IoU) metric, while
ignoring void regions not defined in the ground truth.

\subsection{Automatic Free-Space Mask Generation.}\label{sec:road-mask-generation}
We first evaluated the quality of the automatically
generated masks for free-space, and  conducted ablation experiments to study
how each part of our technique contributes to the algorithm.
Table~\ref{tbl:road-mask-generation} summarizes the results,
in terms of IoU on the Cityscapes dataset.
We emphasize that our model has never seen
the training set ground truth before.
We compare our proposed
superpixel alignment road prior clustering method with directly clustering
the CNN features from the dilated ResNet. As can be seen, superpixel
alignment achieves higher IoU than the raw CNN features.  We can
also see that batch clustering improves both methods, and helps
superpixel alignment to achieve higher IoU.  For the sake of comparison,
we also compare with previous work that combines superpixels and a
saliency map~\cite{Tsutsui_2017_ICCV_Workshops}. We treat the
free-space cluster from the raw CNN features as saliency, and use
these for selecting superpixels. As can be seen, this technique
improves the performance of the raw CNN features, but is still
unable to beat superpixel alignment.

\begin{table}[tb]
\centering
{\small{
\begin{tabular}{lc}
\toprule
    {Technique} & \makecell{IoU} \\
\midrule
  raw CNN features location prior clustering & $0.530$  \\  %
  + batch clustering  & $0.568$   \\
  + superpixel overlap~\cite{Tsutsui_2017_ICCV_Workshops} & $0.620$  \\
    \midrule
  superpixel align location prior clustering & $0.758$  \\
  + batch clustering  & $0.764$   \\
\bottomrule \\
\end{tabular}
}}
\caption{Ablation study results for automatic road mask generation on the Cityscapes training set.}
\label{tbl:road-mask-generation}
\vspace{-3mm}
\end{table}

\begin{figure}[t!]
  \centering
{\small{
\begin{tabular}{c}
  \includegraphics[width=\linewidth]{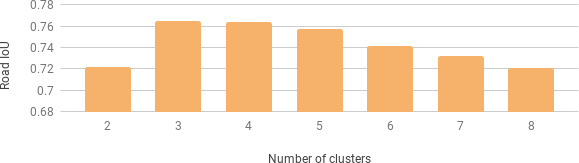} \\
(a) \\
  \includegraphics[width=\linewidth]{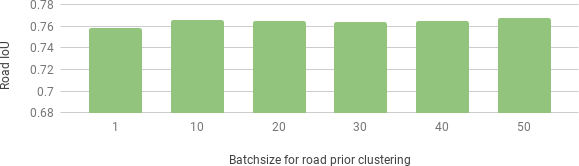} \\
(b) \\
  \includegraphics[width=\linewidth]{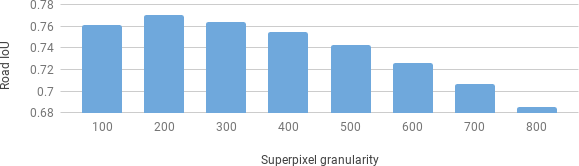} \\
(c) \\
\end{tabular}
}}
 	\caption{Parameter sensitivity with respect to (a) number of clusters, (b) batch size, and (c) superpixel granularity.}
	\label{fig:sensitivities}
\end{figure}

\paragraph{Parameter sensitivity.}
Although our method does not use any annotations, it does rely on some
manually selected parameters. In practice, we chose these values by
visually investigating a small ($\sim10$) number of images. To
measure the sensitivity of our method to these values, we changed each
of three key parameters and compared the final road IoU on the training
set: number of clusters (default 4), batch size (default 30), and
superpixel granularity scale (default 300).
Results are shown in Figure~\ref{fig:sensitivities}.
While performance did vary with differing parameter values, of course,
we found that the final IoU metric differed by only a few percent across
even relatively extreme parameter settings.

In more detail, Figure~\ref{fig:sensitivities}(a) shows the sensitivity for the number
of clusters. We see that having too few clusters makes it
difficult to separate road from other parts of the image, while having
too many also has diminishing returns as the free-space is
eventually  split into multiple clusters.
The effect of
varying the batch size is shown in
Figure~\ref{fig:sensitivities}(b); increasing the
batch size improves the IoU.  Finally, Figure~\ref{fig:sensitivities}(c) shows 
that smaller superpixels tend to work slightly
better, presumably since they avoid undersegmentation which can lead
to false positives (e.g., due to merging free-space with building walls).
These results suggest that our method is
relatively robust to the choice of parameter values.

\subsection{Training a CNN from the Generated Mask}\label{sec:train-cnn}

We next tested our algorithm in the context that it was designed for: automatically
generating pixel-level annotations for training a supervised free-space segmentation
model.
In particular, we used our automatically generated pixel-level annotations from the previous section to train
SegNet~\cite{BadrinarayananSegnet}, although we note that our method is agnostic
to the choice of model so any CNN could be used instead.

\paragraph{Experimental setup.}
We trained SegNet Basic with our generated masks as labels for the
Cityscapes training images using the Chainer
framework~\cite{tokui2015chainer,ChainerCV2017}.  We 
used the validation set to 
evaluate our
method against several baselines, since the test set
annotations are not publicly available (and the evaluation server
restricts the number of submissions to avoid overfitting to the test dataset).
We
emphasize that no hyper-parameters were tuned based on the validation
set; we treated it as if it were the test set.

\paragraph{Baselines.}
We compared our technique with six baseline methods, as summarized in
Table~\ref{tbl:cnn-results}.  The first two baselines serve as simple
indications of how well trivial solutions work on this task:
\textbf{Largest superpixel} uses just the single largest superpixel as
the free-space annotation mask, and \textbf{Bottom half} blindly uses
the bottom half of the image as the free-space mask.  
 In contrast, \textbf{Ground truth as road cluster} uses
the ground truth mask as the clustering results and combines them with
the superpixels in a similar manner as in previous
work~\cite{Tsutsui_2017_ICCV_Workshops}. 
  \textbf{Distant supervision} is the technique of
Tsutsui \textit{et al.}~\cite{Tsutsui_2017_ICCV_Workshops}, which
shares a similar motivation with us and uses external images (which
they call a `distant supervisor') to perform road segmentation in a
weakly supervised manner. 
\textbf{Video segmentation} is the 
technique of Saleh \textit{et al.}~\cite{Saleh_2017_ICCV}, 
which was originally proposed for general background
segmentation and not only uses external images but also videos.  
Finally, \textbf{Fully supervised}
trains the SegNet model from ground truth
annotations.

\begin{table}[tb]
\centering
{\small{
\begin{tabular}{llc}
\toprule
    Method & {Required annotation} & \makecell{IoU} \\
\midrule
  largest superpixel & none & $0.659$  \\
  bottom half & none & $0.720$  \\
  ground truth as road cluster & - & $0.824$  \\
  \midrule
  distant supervision~\cite{Tsutsui_2017_ICCV_Workshops} & {additional images} & $0.800$  \\
  video segmentation~\cite{Saleh_2017_ICCV} & {additional images} & $0.759$   \\
  \midrule
  fully supervised (from~\cite{Tsutsui_2017_ICCV_Workshops})  & {pixel-wise} & $0.853$   \\
  \midrule
   ours (generated masks) & none & $0.761$ \\ %
   ours (CNN training) & none & $0.835$ \\ %
\bottomrule \\
\end{tabular}
}}
\caption{Results evaluated on the Cityscapes validation set.}
\label{tbl:cnn-results}
\end{table}

\paragraph{Evaluation.}
We computed the IoU of SegNet trained on the output of our
weakly-supervised algorithm, and obtained an IoU of $0.835$ on the
Cityscapes validation set.  This is much higher than the trivial
baselines
\textbf{largest superpixel} and \textbf{bottom half}, which
yielded IoUs of 0.659 and 0.720, respectively. 
The relatively high IoU of the
bottom half baseline might make this task seem easy, but we emphasize
that our method has a much lower false positive rate, which is crucial
for employing the method in a practical system to avoid
collisions. In particular, \textbf{Bottom half} gives precision $0.754$,
while our generated masks and trained CNN have a precision of $0.867$
and $0.898$, respectively, thus showing that our method is less prone
to fatal false positives, such as a car being mistaken for road.  
In contrast, false negatives are less important in our application,
since they may just mean that the car is unable to drive to a certain
point, still preserving safe behavior.

Our technique also outperforms
\textbf{distant supervision} and \textbf{video segmentation},
even though they require more annotations.
Of course, our technique also imposes more assumptions, since those
approaches were designed for general video segmentation and our
cues are customized to free-space, but nonetheless we believe
it should be notable because we
do not use any motion cues with video.  
\textbf{Ground truth as road cluster}, 
which can be viewed as an upper bound on the performance of any technique using
superpixels (e.g.,~\cite{Lin2016,Tokmakov16a,Hong2017WeaklySS,
  chen2017no}), yields
an IoU of 0.824.

Of course, \textbf{fully supervised} somewhat outperforms our results (0.853\footnote{This is a
  bit worse than the original SegNet~\cite{BadrinarayananSegnet}
  because we use their simplest model, SegNet Basic, and only train with
  binary classes while the original one used more classes.} vs 0.835).
Nonetheless, it is impressive that our technique achieves $98\%$ of the
IoU of the fully supervised model, 
without requiring the tedious pixel-wise annotations for each image.
This indicates that our proposed method is able to
perform proper free-space segmentation while using no manual
annotations for training the CNN.

\begin{figure*}[t]
  \centering
  \includegraphics[width=170mm]{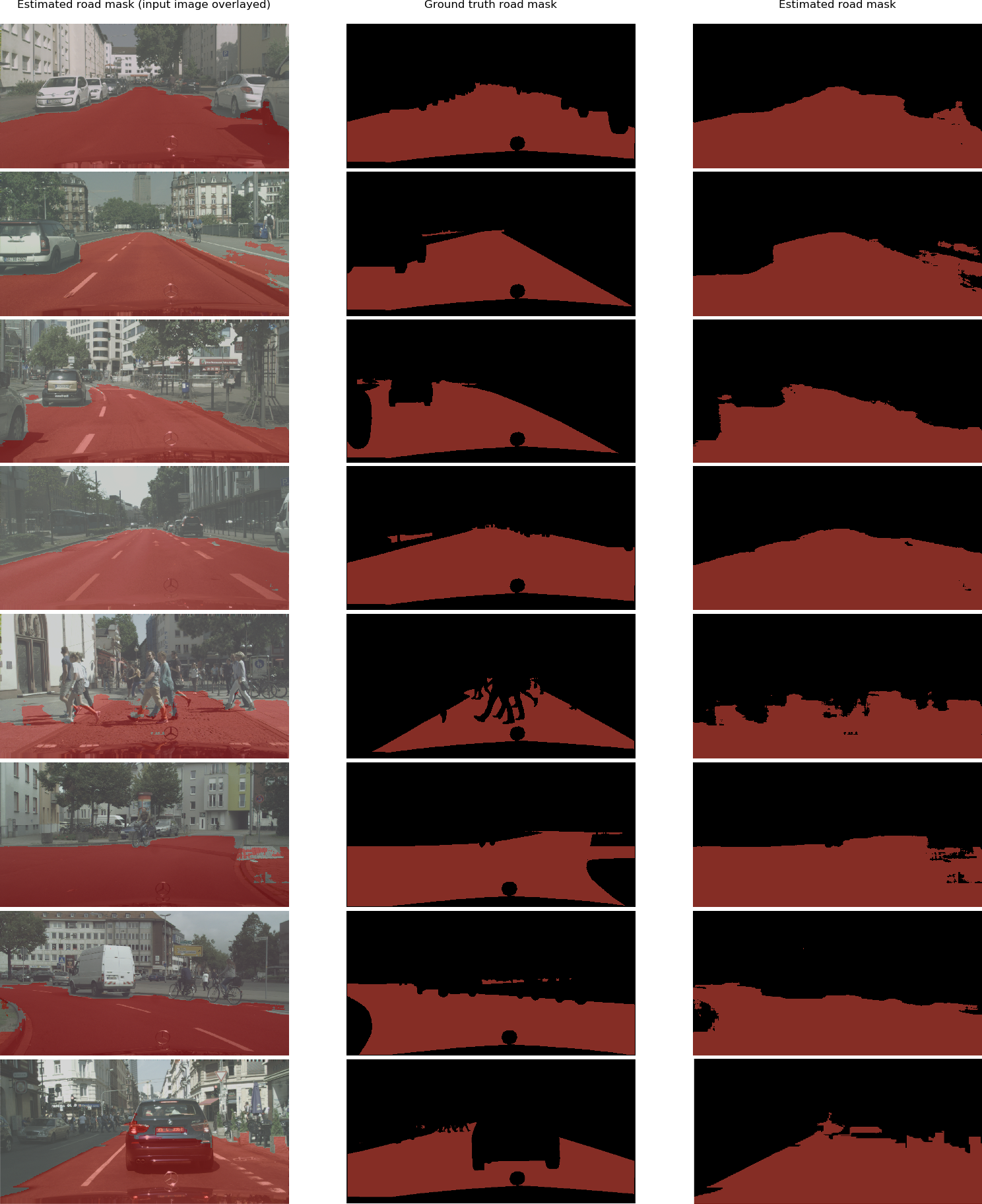}
 	\caption{Sample results on the Cityscapes validation set, comparing our method with ground truth.
 	The last row shows a failure case with a false positive.
 	}
	\label{fig:sample_results}
\end{figure*}

Some sample results on the validation set can be seen in
Figure~\ref{fig:sample_results}.  We see that while our method
typically follows the shape of the true road and avoids labeling cars
as road, it has some trouble with \eg~pedestrian legs and some parts
of the sidewalk being labeled as road. The last row also shows an example of a
false positive, where a car in front of the ego-vehicle is not able to be separated
from the estimated free-space.
In future work, it would be
interesting to investigate more powerful (albeit more
computationally heavy) CNN architectures that might help mitigate these problems~\cite{zhao2016pyramid}.

Finally, we also evaluated our best model on the Cityscapes test set
evaluation server. Results are shown in
Table~\ref{tbl:cnn-results-test-set}, where we can see that consistent
with the validation set results, our method is able to gain better performance than the general video segmentation approach~\cite{Saleh_2017_ICCV}.

\begin{table}[tb]
\centering
{\small{
\begin{tabular}{llc}
\toprule
    {Method} & Required annotation & \makecell{IoU} \\
\midrule
  video segmentation~\cite{Saleh_2017_ICCV}  & {additional images} & $0.785$   \\
   ours (CNN training) & none & $0.857$ \\
\bottomrule \\
\end{tabular}
}}
\caption{Results evaluated on the Cityscapes test set.}
\label{tbl:cnn-results-test-set}
\end{table}

\section{Conclusion}
In this paper, we 
developed a new framework for minimizing human supervision for
free-space segmentation, using assumptions of the
the characteristics of free-space.
 Our method extracts free-space by
performing clustering of superpixel features, which are created by a
novel superpixel alignment method that bases features on the last layer of
an ImageNet-pretrained CNN. We use a location prior to select the
cluster corresponding to free-space and then perform training of a
free-space segmentation CNN.  Unlike previous work, our method needs
no annotations, and experimental results demonstrate superior performance
compared to other methods, even ones that use more information.

As future work, we plan to automatically generate the
location prior conditioned on the input image to better handle 
segmentation of distant free-space, which is a weakness of the current model.
Extending the model to other application domains with high cost of collecting training data, such as robots moving on a house floor
or autonomous water vehicles, is another interesting direction.

\paragraph{Acknowledgments}
We would like to thank the authors of the video segmentation paper~\cite{Saleh_2017_ICCV} for sharing their results on the validation set.
We would also like to thank Masaki Saito and Richard Calland for helpful discussions.

{\small
\bibliographystyle{ieee}
\bibliography{references}
}

\end{document}